\documentclass{article}

\usepackage{PRIMEarxiv}

\usepackage{fancyhdr}       

\usepackage{amsmath,amssymb,amsfonts}
\usepackage{array}
\usepackage[caption=false,font=normalsize,labelfont=sf,textfont=sf]{subfig}
\usepackage{textcomp}
\usepackage{stfloats}
\usepackage{url}
\usepackage{verbatim}
\usepackage{graphicx}
\usepackage{tikz, pgfplots}
\usetikzlibrary{bayesnet}
\usepackage[ruled, lined, linesnumbered, commentsnumbered, longend]{algorithm2e}
\usepackage{booktabs}
\usepackage{multirow}
\usepackage{siunitx}
\usepackage{hyperref}
\usepackage{cite}
\usepackage{comment}
\usepackage{makecell}
\usepackage{wrapfig}

\newcommand{\by}{\mathbf{y}}
\newcommand{\bx}{\mathbf{x}}
\newcommand{\bz}{\mathbf{z}}

\newcommand{\bu}{\mathbf{u}}
\newcommand{\bn}{\mathbf{n}}

\newcommand{\bH}{\mathbf{H}}
\newcommand{\bS}{\mathbf{S}}
\newcommand{\bQ}{\mathbf{Q}}
\newcommand{\bB}{\mathbf{B}}

\newcommand{\I}{\mathbf{I}}

\title{Plug-and-Play split Gibbs sampler: embedding deep generative priors in Bayesian inference
}

\author{
  Florentin Coeurdoux, Nicolas Dobigeon\thanks{This work was supported by the Artificial Natural Intelligence Toulouse Institute (ANITI, ANR-19-PI3A-0004).}  \\
  IRIT/INP-ENSEEIHT \\
  University of Toulouse \\
  Toulouse, France\\
  \texttt{\{Florentin.Coeurdoux, Nicolas.Dobigeon\}@irit.fr} \\
   \And
  Pierre Chainais\thanks{This work was supported by the AI Sherlock Chair (ANR-20-CHIA-0031-01), the ULNE national future investment programme (ANR-16-IDEX-0004) and the Hauts-de-France Region.}  \\
  CNRS, Centrale Lille \\
  University of Lille \\
  Lille, France\\
  \texttt{Pierre.Chainais@centralelille.fr}
}

\begin{document}
\maketitle

\begin{abstract}
This paper introduces a stochastic plug-and-play (PnP) sampling algorithm that leverages  variable splitting to efficiently sample from a posterior distribution. The algorithm based on split Gibbs sampling (SGS) draws inspiration from the alternating direction method of multipliers (ADMM). It divides the challenging task of posterior sampling into two simpler sampling problems. The first problem depends on the likelihood function, while the second is interpreted as a Bayesian denoising problem that can be readily carried out by a deep generative model. Specifically, for an illustrative purpose, the proposed method is implemented in this paper using state-of-the-art diffusion-based generative models. Akin to its deterministic PnP-based counterparts, the proposed method exhibits the great advantage of not requiring an explicit choice of the prior distribution, which is rather encoded into a pre-trained generative model. However, unlike optimization methods (e.g., PnP-ADMM) which generally provide only point estimates, the proposed approach allows conventional Bayesian estimators to be accompanied by confidence intervals at a reasonable additional computational cost. Experiments on commonly studied image processing problems illustrate the efficiency of the proposed sampling strategy. Its performance is compared to recent state-of-the-art optimization and sampling methods.
\end{abstract}


\section{Introduction}
Many scientific problems raise the challenge of inferring an unknown object of interest $\bx \in \mathbb{R}^N$ from partial and noisy measurements $\by \in \mathbb{R}^M$. These inverse problems frequently encountered in image processing are typically formulated as the minimisation task 
\begin{equation}
\underset{\bx}{\min } f(\bx, \by)+ g(\bx)
\label{eq:forward}
\end{equation}
where $f(\cdot,\by)$ denotes the data-fitting term. Due to the ill-posed or ill-conditioned nature of the inverse problem, it is often not possible to uniquely and stably recover $\bx$ from the sole observations $\by$. Therefore, additional information about the unknown object $\bx$ is incorporated in the form of the regularization $g(\cdot)$ to obtain a well-posed estimation problem, leading to meaningful solutions \cite{kaipio2006statistical}. Due to the increasing volume, dimensionality, and variety of available data, solving such inference problems can be computationally demanding and may rely on methods such as variational optimization or stochastic sampling.

Until recently, most of the optimization methods have relied on priors designed as explicit model-based regularizations such as the total variation, promoting piecewise constant behaviors, or the $\ell_1$ norm, promoting sparsity. In this context, convex optimization algorithms have played an important role and their convergence properties have been well-established \cite{chandrasekaran2012convex, repetti2019scalable, chambolle2004algorithm, louchet2013posterior}. However, for an always larger family of problems related to image processing, methods based on explicit convex priors are now significantly outperformed by deep learning based approaches. There exist a number of deep neural network architectures that can directly learn a description of the solution space \cite{dong2014learning, zhang2017beyond, zhang2018ffdnet, schwartz2018deepisp, gregor2010learning}. Such so-called end-to-end approaches that bypass the problem of explicitly defining the prior knowledge  do not even need the knowledge of the forward operator itself. Instead, they are implicitly learnt from a large data set of degraded images (i.e., network input) along with their original versions (i.e., network output) when training the network. However, such end-to-end methods suffer from the lack of interpretability and generality of black-box deep neural networks (DNN). Moreover, they do not take advantage of the generally well-established expertise of the end-users about the acquisition or damaging protocols, which makes the training process particularly energy and data intensive.

To overcome these limitations, more and more deep learning based methods propose to combine DNN with conventional optimization algorithms within the so-called plug-and-play (PnP) framework \cite{venkatakrishnan2013plug}. The main ingredient of PnP approaches is a variable splitting strategy as implemented by half-quadratic splitting (HQS) \cite{geman1995nonlinear} or alternating direction method of multipliers (ADMM) \cite{boyd2011distributed}. The main idea of this splitting consists in introducing an auxiliary variable $\bz$ such that the problem \eqref{eq:forward} rewrites
\begin{equation}
\underset{\bx, \bz}{\min } f(\bx, \by)+g(\bz) \quad \text{subject to } \quad \bx=\bz.
\label{eq:splitpb}
\end{equation}
The equality constraint ensures that solving (\ref{eq:splitpb}) is equivalent to solving the initial problem (\ref{eq:forward}). Adopting an alternate minimization strategy, this tricks permits to separately deal with the data-fitting term and the regularization  \cite{neal2014proximal}. In particular, the subproblem with respect to $\bz$ is solved by using the proximal operator of the regularization term, which can be interpreted as a denoising task. Recent PnP methods replace this proximal mapping by a DNN-based denoiser that implicitly encodes the regularization. They now stand as a reference that yield state-of-the-art performance in a variety of applications \cite{zhang2021plug, ahmad2020plug}.  

However, PnP-based optimization algorithms generally produce point estimates only. More generally, except in special cases \cite{kendall1946advanced}, optimization methods do not give any information about the posterior distribution 
\begin{equation}
    \pi(\bx) \triangleq p(\bx|\by) \propto \exp [-f(\bx,\by)-g(\bx)]
\label{eq:post}
\end{equation}
associated with \eqref{eq:forward} and do not quantify uncertainties. Conversely, Bayesian approaches and Markov chain Monte Carlo (MCMC) methods have the great advantage of providing a comprehensive description of \eqref{eq:post} in very general settings. In particular, this knowledge permits to derive credibility intervals on the parameter $\bx$ of interest. This uncertainty quantification is often of crucial importance, for instance when only very few observations are available \cite{cai2018uncertainty}, when one is interested in extreme events \cite{gaume2010bayesian} or when no ground truth is available, like in astrophysics. There is still a price to pay: sampling methods and MCMC in particular suffer from their high computational cost which can be prohibitive in high-dimensional problems. Optimization-driven Monte Carlo methods \cite{pscp16,duane1987hybrid, pereyra2017maximum} tentatively overcome this limitation. 

More recently, the split Gibbs sampler (SGS) \cite{vono2019split} proposes to sample from an augmented distribution defined as an asymptotically exact data augmentation model \cite{vono2020asymptotically}. By introducing an auxiliary variable as in \eqref{eq:splitpb}, it yields a divide-to-conquer strategy by splitting the initial sampling problem into individual simpler sampling tasks. Sampling according to the augmented distribution with Gibbs steps permits to deal separately with the distinct components of the problem, i.e., the likelihood on the one hand and the prior on the other hand. Per se, SGS can be seen as a stochastic counterpart of HQS or ADMM algorithms. It both makes the sampling more scalable to high dimensions and significantly improves the mixing properties of the Markov chain.

The main contribution of the work reported in this paper is to provide a straightforward and systematic instantiation of the PnP paradigm within a Monte Carlo sampling framework. This is made possible thanks to the splitting strategy implemented by the SGS scheme. Moreover the timeliness of devising such an approach can be easily justified by the recent advances in the design of powerful deep generative models. The proposed approach coined as PnP-SGS is based on three main rationales. First, as any PnP-based methods, PnP-SGS allows Bayesian inference problems to be solved without  explicitly defining a prior distribution, which is rather implicitly encoded into a DNN trained beforehand. Second, we show that diffusion-based or score-based models \cite{im2017denoising, tran2020gan, song2022solving} initially derived for generative purposes can be diverted to be employed as universal stochastic denoisers. Third, PnP-SGS generate samples that can be used to build confidence intervals, which is not possible with its determinisc counterpart, i.e., PnP-ADMM, that only provides point estimates \cite{zhang2021plug, ahmad2020plug}. High dimensional image processing experiments will illustrate the strong potential of the proposed approach when using a diffusion model \cite{dhariwal2021diffusion, ho2020denoising, kingma2021variational, sohl2015deep, song2019generative, song2020improved, song2021scorebased} as a denoiser. These extensive experiments include various inverse problems such as inpainting, super-resolution, and deblurring. The experimental results show that the proposed PnP-SGS is a general approach to solve ill-posed inverse problems in high dimension with superior quality and uncertainty quantification.

Section \ref{sec:def-SGS-PnP} recalls necessary notions about the split Gibbs sampler (SGS) and Denoising Diffusion Probabilistic Models (DDPMs) that will be used as PnP-denoisers in the sequel. Section \ref{sec:appli} describes how the proposed PnP-SGS adapts to several usual inverse problems frequently encountered in image processing. Section \ref{sec:experiments} describes numerical experiments and reports the performances in comparison with state-of-the-art methods. Section \ref{sec:conclusion} finally enlightens the contributions.

\section{SGS and generative models for PnP}
\label{sec:def-SGS-PnP}

\subsection{Split Gibbs sampling (SGS)} \label{subsec:SGS}

Starting from the target posterior distribution (\ref{eq:post}), the introduction of a splitting variable $\bz \in \mathbb{R}^N$ leads to the augmented distribution
\begin{eqnarray}
    \pi_\rho(\bx, \bz) &\triangleq& p\left(\bx, \bz | \by ; \rho^2\right)\label{eq:spdist} \\
    &\propto& \exp \left[-f(\bx,\by)-g(\bz)-\frac{1}{2\rho^2}  \left\| \bx - \bz\right\|_2^2 \right]    
     \nonumber
\end{eqnarray}
where $\rho$ is a positive parameter that controls the coupling between $\bx$ and $\bz$. 
As shown in \cite{vono2020asymptotically}, for a large variety of coupling kernels including the quadratic one, the marginal distribution of $\bx$ under $\pi_\rho$ in (\ref{eq:spdist}) coincides with the target distribution $\pi$ in (\ref{eq:post}) when $\rho^2$ tends to zero, i.e., 
\begin{equation}
\left\|\pi-\pi_\rho\right\|_{\mathrm{TV}} \underset{\rho^2 \rightarrow 0}{\rightarrow} 0
\end{equation}
which defines an asymptotically exact data augmentation scheme \cite{vono2020asymptotically}. In other words, the original target distribution $\pi(\bx)$ in \eqref{eq:post} is recovered from the marginal distribution $\pi_{\rho}(\bx)$ derived from (\ref{eq:spdist}) in the limiting case $\rho \rightarrow 0$. Instead of sampling directly according to $\pi(\bx)$, SGS proposes to sample according to the augmented distribution $\pi_{\rho}(\bx,\bz)$ using Gibbs steps. More specifically, the associated conditional distributions to sample from $\pi_{\rho}(\bx,\bz)$ are given by
\begin{eqnarray}
p\left(\bx \mid \bz, \by; \rho^2\right) & \propto & \exp \left[-f(\bx,\by)-\frac{1}{2\rho^2}  \left\| \bx - \bz\right\|_2^2 \right] \label{eq:condx}\\
p\left(\bz \mid \bx ; \rho^2\right) & \propto & \exp \left[-g(\bz)-\frac{1}{2\rho^2}  \left\| \bz - \bx\right\|_2^2 \right]   \label{eq:condz}
\end{eqnarray}
It is now clear that sampling alternatively from \eqref{eq:condx} and \eqref{eq:condz} dissociates the potential functions $f(\cdot,\by)$ and $g(\cdot)$ associated with the likelihood and the prior distribution, respectively. As a consequence, SGS inherits from well-known advantages already exhibited by its deterministic counterparts (i.e., HGQ and ADMM), e.g., easier implementations, faster convergences and possibly distributed computations. In particular, sampling according to \eqref{eq:condx} can be interpreted as solving the initial problem defined by the same potential function $f(\cdot,\by)$ but now granted with a Gaussian prior distribution of mean $\bz$ and diagonal covariance matrix $\rho^2 \I$. It is thus expected to be significantly simpler than sampling according to the initial posterior distribution $\pi(\bx)$ defined by \eqref{eq:post}. 

Moreover, it is worth noting that the conditional distribution \eqref{eq:condz} can be interpreted as the posterior distribution associated with a Bayesian denoising problem. Its goal is to recover an object $\bz$ from a noisy observations $\bx$ contaminated by an additive white Gaussian noise with variance $\rho^2$. Instead of sampling directly from \eqref{eq:condz}, we propose to resort to deep generative models used as stochastic denoisers. Generative adversarial network (GAN), variational autoencoders (VAE) or more recently denoising diffusion probabilistic models (DDPM) are powerful candidates to tackle this task \cite{im2017denoising, tran2020gan, song2022solving}. Due to the high interest they have received recently, this work instantiates the PnP-SGS framework and reports experimental results based on DDPM-based denoisers.  Note however that any pre-trained probabilistic denoising generative model can be plugged into the proposed approach. 

\subsection{Denoising diffusion probabilistic models (DDPM)} \label{sec:denoising}

Denoising diffusion models \cite{dhariwal2021diffusion, ho2020denoising, kingma2021variational, sohl2015deep} and score based models \cite{song2019generative, song2020improved, song2021scorebased} are trendy classes of generative models. They have recently drawn significant attention from the community due to their state-of-the-art performances. Although nourished by different inspirations, they share very similar aspects and can be presented as variants of each other \cite{huang2021variational, kingma2021variational, song2021scorebased}. They are often referred to under the generic name {\em diffusion models}.\\

\subsubsection{DDPM as generative models} A denoising diffusion probabilistic model \cite{dhariwal2021diffusion} makes use of two Markov chains: a forward chain that perturbs data to pure noise, and a backward chain that converts noise back to data. The former is typically model-based designed with the goal to transform any data distribution into a simple prior distribution, i.e., a standard Gaussian. Conversely the latter Markov chain aims at  reversing the noising process by learning transition kernels parameterized by a DNN. Once the DNN has beeen trained,  new data points can be generated   by first drawing from the prior distribution, and then sampling through the backward Markov chain.

Formally, given a data distribution $\bu_0 \sim p\left(\bu_0\right)$, the forward Markov process generates a sequence of random variables $\bu_t \in \mathbb{R}^N$, $t \in\left\{0,\ldots, T\right\}$ according to the transition kernel $p\left(\bu_t \mid \bu_{t-1}\right)$. Using the probability chain rule and the Markovian property, the joint distribution $p\left(\bu_1, \ldots, \bu_T \mid \bu_0\right)$ can be factorized as 
\begin{equation}
p\left(\bu_1, \ldots, \bu_T \mid \bu_0\right)=\prod_{t=1}^T p\left(\bu_t \mid \bu_{t-1}\right) .
\label{eq:diff}
\end{equation}
In DDPMs, the transition kernel $p\left(\bu_t \mid \bu_{t-1}\right)$ is arbitrarily chosen to incrementally transform the data distribution $p\left(\bu_0\right)$ into a tractable prior distribution $p(\bu_{T})\approx \mathcal{N}\left(\bu_T ; \boldsymbol{0}, \I \right)$. One typical design for the transition kernel exploits a Gaussian perturbation and the most common choice for the transition kernel is
\begin{equation}
p\left(\bu_t \mid \bu_{t-1}\right)=\mathcal{N}\left(\bu_t ; \sqrt{1-\beta(t)} \bu_{t-1}, \beta(t) \mathbf{I}\right)
\label{eq:forw}
\end{equation}
where $\beta(t) \in(0,1)$ is a predefined function which plays a key role. It directly adjusts the amount of noise along the process such that larger values lead to noisier samples. Conventionally, it is chosen as a linearly increasing function \cite{ho2020denoising}. More recent techniques have proposed to use cosine-based functions \cite{nichol2021improved}.
Intuitively speaking, this forward process slowly injects noise into data until all structures are lost and only noise prevails. 

For generating new data samples, DDPMs start by first drawing a sample $\bu_T$ from an instrumental prior distribution $q\left(\bu_T\right)=\mathcal{N}\left(\bu_T ; \mathbf{0}, \I\right)$. Then DDPMs gradually remove noise by running a Markov chain in the reverse time direction. This Markov chain is defined thanks to a kernel modeled by DNNs. The learnable transition kernel $q_{\boldsymbol{\theta}}\left(\bu_{t-1}| \bu_t\right)$ takes the form of
\begin{equation}
q_{\boldsymbol{\theta}}\left(\bu_{t-1} \mid \bu_t\right)=\mathcal{N}\left(\bu_{t-1} ; \boldsymbol{\mu}_{\boldsymbol{\theta}}\left(\bu_t, t\right), \boldsymbol{\Sigma}_{\boldsymbol{\theta}}\left(\bu_t, t\right)\right)
\label{eq:rev}
\end{equation}
where the mean $\boldsymbol{\mu}_{\boldsymbol{\theta}}\left(\bu_t, t\right)$ and the covariance matrix $\boldsymbol{\Sigma}_{\boldsymbol{\theta}}\left(\bu_t, t\right)$ are DNNs parametrized by $\boldsymbol{\theta}$ and $t$ with $\bu_t$ as an input.

\subsubsection{DDPM as stochastic denoisers} \label{subsec:denoising}
According to the above discussion, it is clear that the forward diffusion process \eqref{eq:forw} progressively adds noise to a noise-free image $\bu_0$. Following a discretization scheme generally adopted by these deep generative models, each $\bu_t$ corresponds to a scaled version of $\bu_{t-1}$ corrupted by a Gaussian noise with covariance matrix $\beta(t) \I$. Thanks to the factorization induced by the direct Markov chain and the Gaussian nature of the transition kernel, the transition from the original image $\bu_0$ to any intermediate noisy image $\bu_t$ can be written as
\begin{eqnarray}
     p\left(\bu_t \mid \bu_0\right) & = & \mathcal{N}\left(\bu_t ; \sqrt{\bar{\alpha}(t)} \bu_0,{\alpha}(t) \I \right)     \label{eq:bigforward} \\
    \text{where } \quad 
    \alpha(t) & = & \prod_{j=1}^t\left(1-\beta(j)\right)  \label{eq:def_alpha}
\end{eqnarray}
and $\bar{\alpha}(t) = 1-{\alpha}(t)$. In other words, at any arbitrary time instant ${t^{*}} < T$, the image $\bu_{{t^{*}}}$ resulting from ${t^{*}}$ steps of the forward process is a noisy version of the input image $\bu_0$ corrupted by a Gaussian noise of variance $\alpha({t^{*}})$. 

Therefore, it appears that a trained DDPM can be used as a stochastic Gaussian denoiser. Contrary to the normal use of a DDPM as a generator (see above), the key idea is rather to start the backward diffusion process from a noisy image $\bu_{{t^{*}}}$ for some ${t^{*}}$ and not as usual from a realization of noise $\bu_T$. The noise-free image $\bu_0$ can be recovered by applying the backward process defined by \eqref{eq:rev} from time instant ${t^{*}}$. 

\subsection{Proposed DDPM-based PnP-SGS algorithm}
In a nutshell, the proposed PnP-SGS alternatively samples according to the conditional posterior distributions \eqref{eq:condx} and \eqref{eq:condz}. Along this iterative process, SGS generates a set of $N_{\mathrm{MC}}$ samples $\left\{\bx^{(n)},\bz^{(n)}\right\}_{n=1}^{N_{\mathrm{MC}}}$ asymptotically distributed according to the augmented posterior $\pi_{\rho}(\bx,\bz)$. From this set of samples, various Bayesian quantities can be approximated, such as Bayesian estimators and credibility intervals. In particular, the samples $\left\{\bx^{(n)}\right\}_{n=1}^{N_{\mathrm{MC}}}$ are marginally distributed according to $\pi_{\rho}(\bx)$. Thus the minimum mean square estimator (MMSE or posterior mean) $\hat{\bx}_{\textrm{MMSE}}=\mathrm{E}[\bx|\by]$ associated with $\pi_{\rho}$ can be easily approximated by the empirical average
\begin{equation}
    \hat{\bx}_{\textrm{MMSE}} \approx \frac{1}{N_{\mathrm{MC}}-N_{\mathrm{bi}}} \sum_{n=N_{\mathrm{bi}}+1}^{N_{\mathrm{MC}}} \bx^{(n)}
\end{equation}
where $N_{\mathrm{bi}}$ is the number of burn-in iterations.

Regarding the first step of SGS, sampling according to \eqref{eq:condx} is problem dependent and should be suitably adapted to the targeted task. For illustration purpose, it will be explicitly specified for various imaging problems in Section \ref{sec:appli}. As expected and already pointed out in Section \ref{subsec:SGS}, it will be shown that sampling according to \eqref{eq:condx} is significantly simpler than directly sampling according to the target posterior distribution $\pi(\bx)$ defined by \eqref{eq:post}.

Regarding the second step of SGS, at the $n$th iteration of the algorithm, sampling according to \eqref{eq:condz} is interpreted as a stochastic denoising of the current value $\bx^{(n)}$. This sampling according to \eqref{eq:condz} is performed in a PnP manner thanks to a previously trained DDPM, following  the strategy detailed in Section \ref{subsec:denoising}. With the notations adopted in the previous paragraph, it assigns the current sample $\bx^{(n)}$ to the variable $\bu_{{t^{*}}}$ for some ${t^{*}}$ at iteration $n$ and then iterates the backward diffusion \eqref{eq:rev}. After ${t^{*}}$ steps, the produced denoised image $\bu_0$ is allocated to the new sample $\bz^{(n)}$ according to \eqref{eq:condz} of the current SGS iteration. Note that DDPMs used as generators are known to be generally computationally demanding due to the number $T$ of overall steps involved in the backward process. The proposed approach obviates this impediment by initiating the process from a generally weakly noisy image, which significantly reduces the necessary number ${t^{*}} \ll T$ of denoising steps to be applied \cite{chung2022come}. Next section provides some insights into this number ${t^{*}}$ and proposes a systematic and reliable strategy to adjust it.

\subsection{Some insights into the number ${t^{*}}$ of backward steps }\label{subsec:insights_t}
This section discusses the role and the tuning of the time instant ${t^{*}}$ which defines the number of denoising steps to be applied at a given iteration of the SGS sampler. As already stated, Eq. \eqref{eq:bigforward} shows that the variance of the noise corrupting $\bu_0$ after $t^*$ transitions of the forward Markov chain is $\alpha(t^*)$. This variance is defined by the product \eqref{eq:def_alpha} of continuous strictly monotone functions $\beta(\cdot)$, thus it is also continuous and strictly monotone. This has two consequences: \emph{i)} a level of noise  $\alpha({t^{*}})$ is associated to a unique instant ${t^{*}}$ of the forward diffusion process (i.e., $\alpha(t)$ is an invertible function of $t$) and $\emph{ii)}$ the larger $t^*$, the noisier the image $\bu_{t^*}$. Reciprocally, when applying the backward diffusion to a noisy image, the larger $t^*$, the higher the impact of the denoising, that is of the regularization. 
Note that the DDPM, that is used for regularization here, has no explicit hyperparameter.
An important consequence is that, within the framework of PnP-SGS, the number $t^*$ of denoising steps can be interpreted as the hyperparameter that adjusts the amount of imposed regularization, the coupling parameter $\rho$ being kept fixed. 

The proposed approach capitalizes on the explicit and unequivocal mapping between the hyperparameter $t^{*}$ and the  variance $\alpha(t^*)$ of the noise contained in $\bu_{t^*}$. This relationship permits a simple and efficient strategy to set the number $t^*$ of required denoising steps \eqref{eq:rev} when sampling according to \eqref{eq:condz}. Given a current sample $\bx^{(n)}$ generated by SGS, the identification of the appropriate instant $t^*$ to generate $\bz^{(n)}$ according to \eqref{eq:condz} boils down to estimating the level $\alpha(t^*)$ of the noise corrupting the sample $\bx^{(n)}$. This is possible using any good conventional estimator $\hat{\sigma}=\Phi(\bx^{(n)})$ of the noise level in $\bx^{(n)}$ \cite{donoho1994ideal, guo2021gaussian, li2022single}, see Appendix \ref{sec:denoise_exp} for implementation details. Since the function $t \rightarrow \alpha(t)$ is invertible, one can finally set %
$\widehat{t^{*}} = \alpha^{-1}(\widehat{\sigma}^2)$ to start the backward diffusion \eqref{eq:rev}. Appendix \ref{app:start_time} discusses technical details of the inversion of $\alpha(\cdot)$. 

In practice, during the experiments reported in Section \ref{sec:experiments}, the  number $\widehat{t^*}$ of achieved steps has been shown to stabilize at a fixed value after the burn-in period of PnP-SGS. Therefore the transition kernel associated with the denoising procedure becomes invariant, which ensures that SGS converges towards a stationary distribution $\pi_{\rho}$; recall that $\rho$ is fixed, typically of order 1, see Appendix \ref{sec:implem}.
The resulting distribution $\pi_\rho$ is eventually similar to \eqref{eq:spdist} where the role of the explicit regularizing potential $g(\cdot)$ has been implicitly replaced by the DDPM.

Algorithm~\ref{alg:PnP-SGS} describes the final sampling PnP-SGS algorithm using a DDPM for the denoising step, with the proposed strategy to set the hyperparameter $t^*$. 

\SetKw{KwDownTo}{downto}
\begin{algorithm}
\caption{PnP-SGS using DDPM}\label{alg:PnP-SGS}
\SetKwInOut{KwIn}{Input}
\SetKwInOut{KwOut}{Output}

\KwIn{Parameter $\rho^2$, total number of iterations $N_{\mathrm{MC}}$, number of burn-in iterations $N_{\mathrm{bi}}$, pre-trainted DDPM $s_{\boldsymbol{\theta}}(\cdot, \cdot)$, scheduling variance function $\alpha(\cdot)$, initialization $\bz^{(0)}$}

\For{$n \gets 1$ \KwTo $N_{\mathrm{MC}}$}{
    \# \small{\it Sampling the variable of interest $\bx^{(n)}$} \\
    Draw $\bx^{(n)} \sim p(\bx \mid \bz, \by; \rho^2)$ according to (\ref{eq:condx}) \\
    \# \small{\it Estimating noise level in $\bx^{(n)}$} \\
    Set $\hat{\sigma} = \Phi(\bx^{(n)})$ using \cite{donoho1994ideal} \\
    \# \small{\it Setting the number of diffusion steps to denoise $\bx^{(n)}$} \\
    Set ${\widehat{t^*}} = \alpha^{-1}(\hat{\sigma}^2)$ \\
    \# \small{\it Sampling the splitting variable $\bz^{(n)}$ according to (\ref{eq:condz})} \\
    Set $\bu_{{\widehat{t^*}}} = \bx^{(n)}$ \\
    \For{$j \gets {\widehat{t^*}}$ \KwDownTo $1$}{	
        Draw $\bu_{j-1} \sim q_{\boldsymbol{\theta}}(\bu_{j-1} \mid \bu_{j})$ according to \eqref{eq:rev}
    }
    Set $\bz^{(n)} = \bu_0$
}
\KwOut{Collection of samples $\left\{\bx^{(n)}, \bz^{(n)}\right\}_{t=N_{\mathrm{bi}+1}}^{N_\mathrm{MC}}$ asymptotically distributed according to (\ref{eq:spdist}).}
\end{algorithm}

\section{Application to Bayesian inverse problems}
\label{sec:appli}
The proposed PnP-SGS method is now instanciated for three different imaging problems, namely deblurring, inpainting and superresolution, following the protocols already considered in \cite{vono2019split}. The considered linear Gaussian inverse problems define an archetypal class of problems that can efficiently tackled by the proposed method. More specifically, a degraded image $\by$ is observed and one wants to infer a restored image $\bx$ under the linear model
\begin{equation}
\by=\bH \bx+ \bn
\label{eq:linear}
\end{equation}
where $\bH$ is a forward operator and $\bn$ accounts for noise or error modeling. Assuming that $\bn$ is a Gaussian random vector with covariance matrix $\boldsymbol{\Omega}^{-1}$, the likelihood function associated with the observation $\by$ writes
$$
p(\by \mid \bx) \propto \exp \left[-\frac{1}{2}(\bH \bx-\by)^T \boldsymbol{\Omega}(\bH \bx-\by)\right] .
$$
In most applicative contexts, inferring the unknown parameter vector $\bx$ from the observation vector $\by$ under the linear model (\ref{eq:linear}) is known to be an ill-posed or ill-conditioned inverse problem. A common approach to tackle such problems consists in using some regularization defined through the choice of a prior distributon $p(\bx) \propto \exp \left[-g(\bx)\right]$, leading to the posterior distribution \eqref{eq:post}. Instead of explicitly specifying the potential function $g(\cdot)$ in \eqref{eq:post}, the proposed PnP-SGS algorithm targets an augmented posterior similar to \eqref{eq:spdist} to capitalize on a pre-trained denoising diffusion model presented in Section \ref{sec:denoising}. 

The three considered tasks mainly differ by the nature of the linear operator $\bH$. Following the SGS algorithmic scheme, a special care should be taken to ensure an efficient sampling according to the conditional posterior \eqref{eq:condx} which involves $\bH$, see Algo. \ref{alg:PnP-SGS}, line 3. Since the sampling according to \eqref{eq:condz} does not depend on the forward operator, it is achieved in a unique manner from a DDPM. Thus the sequel of this section is only devoted to the technical derivations associated with \eqref{eq:condx}. Experimental results obtained by the proposed PnP-SGS will be reported in Section \ref{sec:experiments}.

\subsection{Image deblurring}\label{subsec:deblurring}
In this setup, the operator $\bH$ is assumed to be an $N \times N$ circulant convolution matrix associated to a blurring kernel. The noise covariance matrix is assumed to be diagonal, i.e., $\boldsymbol{\Omega}^{-1}=\operatorname{diag}\left[\sigma_1^2, \ldots, \sigma_N^2\right]$ where distinct diagonal elements mimic a spatially-variant noise level. Even when choosing a simple model-based regularizing potential $g(\cdot)$, direct sampling according to the posterior distribution (\ref{eq:post}) may remain a challenging task, mainly due to the presence of the precision matrix $\boldsymbol{\Omega}$ which prevents a direct computation in the Fourier domain. Conversely, the proposed PnP-SGS algorithm yields the conditional distribution (\ref{eq:condx}) defined here as
\begin{equation}
    p(\bx \mid \bz,\by ; \rho^2)=\mathcal{N}\left(\bx;\boldsymbol{\mu}_{\bx}, \bQ_{\bx}^{-1} \right)
    \label{eq:gausx}
\end{equation}
with
\begin{equation}
\left\{\begin{array}{l}
\bQ_{\bx}= \bH^T \boldsymbol{\Omega} \bH+ \frac{1}{\rho^{2}} \mathbf{I}_N \\
\boldsymbol{\mu}_{\bx} = \bQ_{\bx}^{-1} \left(\bH^T \boldsymbol{\Omega}\by + \frac{1}{\rho^2} \bz\right).
\end{array}\right.
\label{eq:mu_sig}
\end{equation}
Thanks to the splitting trick inherent to the proposed PnP-SGS algorithm, this step does not depend on $g(\cdot)$ and boils down to a high-dimensional Gaussian sampling task. This task has been deeply investigated in \cite{Vono_SIREV_2022} and can be efficiently achieved by using the auxiliary method of \cite{marnissi2018aux}. Finally, sampling from (\ref{eq:condz}) is straightforward using the pre-trained network as discussed in Section \ref{sec:denoising}.

\subsection{Image inpainting}\label{subsec:inpainting}
Image inpainting problems aim at recovering an original image $\bx \in \mathbb{R}^N$ from the noisy and partial measurements $\by \in \mathbb{R}^M$ under the linear model (\ref{eq:linear}). The operator $\bH \in \{0,1\}^{N\times M}$ now stands for a binary matrix associated with a irregular subsampling with $M \ll N$. The noise is assumed to be white and Gaussian such that $\boldsymbol{\Omega}^{-1}=\sigma^2 \I_M$. As for the deblurring task, the conditional distribution \eqref{eq:condx} is (\ref{eq:gausx}) with 
\begin{equation}
\left\{\begin{array}{l}
\bQ_{\bx}= \frac{1}{\sigma^{2}} \bH^T \bH+ \frac{1}{\rho^{2}} \I_N \\
\boldsymbol{\mu}_{\bx} = \bQ_{\bx}^{-1} \left(\frac{1}{\sigma^2}\bH^T \by + \frac{1}{\rho^2} \bz\right).
\end{array}\right.
\label{eq:mu_sig_inpainting}
\end{equation}
The difficulty of sampling according to this Gaussian distribution comes from the operator $\bH$ which is not diagonalizable in the Fourier domain. However, since it consists of a subset of rows of the identity matrix $\I_N$, one has $\bH \bH^T=\I_M$ and the Sherman-Morrison-Woodbury formula yields
\begin{equation}
\bQ_{\bx}^{-1}=\rho^2\left(\I_N-\frac{\rho^2}{\sigma^2+\rho^2} \bH^T \bH\right) .    
\label{eq:SMW}
\end{equation}
Since $\bH^T\bH$ is diagonal, the covariance matrix \eqref{eq:SMW} is diagonal and sampling from \eqref{eq:gausx} can be conducted efficiently with the exact perturbation-optimization (E-PO) algorithm \cite{marnissi2018aux}.

\subsection{Image super-resolution}
Image super-resolution is characterized by a forward model composed of a blurring kernel followed by a subsampling step. The forward operator writes
\begin{equation}
    \bH = \bS\bB
\label{eq:SR}
\end{equation}
where $\bB$ is a $N \times N$ circulant convolution matrix, as in Section \ref{subsec:deblurring}, and $\bS \in \{0,1\}^{M \times N}$ is associated with a binary mask, as in Section \ref{subsec:inpainting}. The noise $\bn$ is assumed to be white and Gaussian. To fully benefit from the advantages of the SGS, two auxiliary variables $\bz_1$ and $\bz_2$ are introduced to define the augmented posterior distribution
\begin{equation}
 p(\bx,\bz_1,\bz_2 | \by; \rho_1^2,\rho_2^2) \propto \exp \left[ - \frac{1}{2 \sigma^2} \|\by- \bS\bz_1\|^2 - \frac{1}{2\rho_{1}^{2}} \|\bz_1-\bB\bx\|^2 -  g(\bz_2) - \frac{1}{2\rho_{2}^{2}} \|\bz_2-\bz_1\|^2 \right]
\end{equation}
This double splitting leads to a SGS algorithm which samples alternatively according to the conditional distributions
\begin{align}
    & p(\bz_1 \mid \bx, \by) \propto \exp \left[ - \frac{1}{2 \sigma^2} \|\by- \bS\bz_1\|^2 - \frac{1}{2\rho_{1}^{2}} \|\bz_1-\bB\bx\|^2 \right] \label{eq:SRin} \\
    & p(\bx \mid \bz_1, \bz_2) \propto \exp \left[ - \frac{1}{2\rho^2_1} \|\bz_1 - \bB\bx\|^2 - \frac{1}{2\rho_{2}^{2}} \|\bz_1-\bz_2\|^2 \right] \label{eq:SRdeb} \\
    & p( \bz_2 \mid \bz_1 ) \propto \exp \left[ - g(\bz_2) - \frac{1}{2\rho_{2}^{2}} \|\bz_2-\bz_1\|^2  \right] \label{eq:SRdz}
\end{align}
 
The two distributions (\ref{eq:SRin}) and (\ref{eq:SRdeb}) correspond to the previously discussed tasks of inpainting and deblurring, respectively. Sampling according to the last one \eqref{eq:SRdz} is achieved thanks to a DDPM used as a stochastic PnP denoiser.

\section{Experiments} \label{sec:expe}
\label{sec:experiments}

\subsection{Experimental setup}

Experiments have been conducted on two data sets each composed of 1000 RGB images of size $256\times 256$ with various characteristics, namely FFHQ $256 \times 256$ \cite{karras2019style}, and Imagenet $256 \times 256$ \cite{deng2009imagenet}. Pre-trained diffusion models have been directly taken from \cite{dhariwal2021diffusion, choi2021ilvr} and used  without any additional fine-tuning. The test images have never been seen by the model while training to avoid any bias due to potentially over-fitted pre-trained models. All images are normalized to the range $(0, 1)$. For the inversion tasks described in Section \ref{sec:appli}, the forward measurement operators have been designed as follows:
\begin{itemize}
    \item \emph{deblurring}: two blurring kernels are considered, namely a Gaussian blur with a kernel size of $61 \times 61$ with standard deviation of $3.0$, and a randomly generated motion blur\footnote{Following the code available at \href{https://github.com/LeviBorodenko/motionblur}{code}.} with size $61 \times 61$ and intensity value $0.5$,
    \item \emph{inpainting}: $80\%$ of the total pixels have been randomly masked accross all RGB channels,
    \item \emph{superresolution}: the operator $\bS$ corresponds to a downsampling factor $d=4$ in both directions and the operator $\bB$ stands for a Gaussian blur with a kernel size of $9 \times 9$ and a standard deviation of $1.5$. 
\end{itemize}

\renewcommand\arraystretch{1.2}
\begin{table*}
\scalebox{0.92}{%
\centering
\begin{tabular}{ccccccccc}
\toprule
 & & PnP-SGS & SPA \cite{vono2019split} & TV-ADMM & PnP-ADMM \cite{chan2016plug}  &Score-SDE \cite{song2021scorebased} & DDRM \cite{kawar2022denoising} & MCG \cite{chung2022improving} \\
 \hline
 \multirow{4}{*}{\rotatebox{90}{{Inpainting}}} 
   &  {PSNR $\uparrow$} & $\mathbf{32.59}$ & 26.09 & 22.03 & 8.41 & 13.52 & 9.19 & \underline{21.57}\\ 
   &  {SSIM $\uparrow$} & $\mathbf{0.913}$ & 0.524 & 0.784 & 0.325 & 0.437 & 0.319 & \underline{0.751}\\ 
   &  {FID $\downarrow$} & \underline{37.36} & 71.12 & 181.56 & 123.61 & 76.54 & \underline{69.71} & $\mathbf{29.26}$\\ 
   &  {LPIPS $\downarrow$} & $\mathbf{0.144}$ & 0.785 & 0.463 & 0.692 & 0.612 & 0.587 & \underline{0.286}\\ 
   \hline 
   \multirow{4}{*}{\rotatebox{90}{\makecell{Deblurring\\ (Gaussian)}}} 
   &  {PSNR $\uparrow$} & $\mathbf{27.96}$ & 23.17 & 22.37 & \underline{24.93} & 7.12 & 23.36 & 6.72\\ 
   &  {SSIM $\uparrow$} & $\mathbf{0.837}$ & 0.499 & \underline{0.801} & \underline{0.812} & 0.109 & 0.767 & 0.051\\ 
   &  {FID $\downarrow$} & $\mathbf{59.667}$ & 78.67 & 186.74 & 90.42 & 109.07 & \underline{74.92} & 101.2\\ 
   &  {LPIPS $\downarrow$} & $\mathbf{0.331}$ & 0.452 & 0.507 & 0.441 & 0.403 & \underline{0.332} & 0.340\\ 
   \hline 
   \multirow{4}{*}{\rotatebox{90}{\makecell{Deblurring\\ (motion)}}} 
   &  {PSNR $\uparrow$} & $\mathbf{28.46}$ & 17.73 & 21.36 & \underline{24.65} & 6.58 & N/A & 6.72\\ 
   &  {SSIM $\uparrow$} & $\mathbf{0.828}$ & 0.211 & 0.751 & \underline{0.825} & 0.102 & N/A & 0.055\\ 
   &  {FID $\downarrow$} & $\mathbf{60.01}$ & 103.87 & 152.39 & \underline{89.08} & 292.28 & N/A & 310.5\\ 
   &  {LPIPS $\downarrow$} & $\mathbf{0.294}$ & 0.446 & 0.508 & \underline{0.405} & 0.657 & N/A & 0.702\\  
   \hline 
   \multirow{4}{*}{\rotatebox{90}{\makecell{Superres. \\ ($\times 4$)}}} 
   &  {PSNR $\uparrow$} & $\mathbf{25.99}$ & N/A & 23.86 & \underline{26.55} & 17.62 & 25.36 & 19.97\\ 
   &  {SSIM $\uparrow$} & \underline{0.812} & N/A & 0.803 & $\mathbf{0.865}$ & 0.617 & 0.835 & 0.703\\ 
   &  {FID $\downarrow$} & $\mathbf{58.82}$ & N/A & 110.64 & 66.52 & 96.72 & \underline{62.15} & 87.64\\ 
   &  {LPIPS $\downarrow$} & $\mathbf{0.279}$ & N/A & 0.428 & 0.353 & 0.563 & \underline{0.294} & 0.520\\       
   \bottomrule
\end{tabular}
}
\caption{ FFHQ $256\times256$ data set: image reconstruction (PSNR, SSIM) obtained by the compared methods. \textbf{Bold}: best, \underline{underline}: second.\label{tab:FFHQ-metrics}}
\end{table*}

\begin{table*}
\scalebox{0.92}{%
\centering
\begin{tabular}{ccccccccc}
\toprule
 & & PnP-SGS & SPA \cite{vono2019split} & TV-ADMM & PnP-ADMM \cite{chan2016plug}  &Score-SDE \cite{song2021scorebased} & DDRM \cite{kawar2022denoising} & MCG \cite{chung2022improving} \\
 \hline
 \multirow{4}{*}{\rotatebox{90}{{Inpainting}}} 
   &  {PSNR $\uparrow$} & $\mathbf{25.22}$ & \underline{23.14} & 20.96 & 8.39 & 18.62 & 14.29 & 19.03\\ 
   &  {SSIM $\uparrow$} & $\mathbf{0.870}$ & 0.802 & 0.676 & 0.300 & 0.517 & 0.403 & 0.546\\ 
   &  {FID $\downarrow$} & $\mathbf{34.28}$ & 41.33 & 189.3 & 114.7 & 127.1 & 114.9 & \underline{39.19}\\ 
   &  {LPIPS $\downarrow$} & $\mathbf{0.297}$ & \underline{0.323} & 0.510 & 0.677 & 0.659 & 0.665 & 0.414\\ 
   \hline 
   \multirow{4}{*}{\rotatebox{90}{\makecell{Deblurring\\ (Gaussian)}}} 
   &  {PSNR $\uparrow$} & \underline{21.76} & 21.08 & 19.99 & $\mathbf{21.81}$ & 15.97 & 22.73 & 16.32\\ 
   &  {SSIM $\uparrow$} & \underline{0.701} & 0.577 & 0.634 & 0.669 & 0.436 & $\mathbf{0.705}$ & 0.441\\ 
   &  {FID $\downarrow$} & $\underline{64.12}$ & 98.78 & 155.7 & 100.6 & 120.3 & $\mathbf{63.02}$ & 95.04\\ 
   &  {LPIPS $\downarrow$} & $\mathbf{0.399}$ & 0.537 & 0.588 & 0.519 & 0.667 & \underline{0.427} & 0.550\\ 
   \hline 
   \multirow{4}{*}{\rotatebox{90}{\makecell{Deblurring\\ (motion)}}} 
   &  {PSNR $\uparrow$} & \underline{21.47} & 20.49 & 20.79 & $\mathbf{21.98}$ & 7.21 & N/A & 5.89\\ 
   &  {SSIM $\uparrow$} & \underline{0.695} & 0.681 & 0.677 & $\mathbf{0.702}$ & 0.120 & N/A & 0.037\\ 
   &  {FID $\downarrow$} & $\mathbf{47.57}$ & 91.51 & 138.8 & 89.76 & 98.25 & N/A & 186.9\\ 
   &  {LPIPS $\downarrow$} & $\mathbf{0.372}$ & 0.538 & 0.525 & 0.483 & 0.591 & N/A & 0.758\\  
   \hline 
   \multirow{4}{*}{\rotatebox{90}{\makecell{Superres.\\ ($\times 4$)}}} 
   &  {PSNR $\uparrow$} & \underline{24.33} & N/A & 22.17 & 23.75 & 12.25 & $\mathbf{24.96}$ & 13.39\\ 
   &  {SSIM $\uparrow$} & \underline{0.772} & N/A & 0.679 & 0.761 & 0.256 & $\mathbf{0.790}$ & 0.227\\ 
   &  {FID $\downarrow$} & $\mathbf{59.09}$ & N/A & 130.9 & 97.27 & 170.7 & 59.57 & 144.5\\ 
   &  {LPIPS $\downarrow$} & $\mathbf{0.418}$ & N/A & 0.523 & 0.433 & 0.701 & 0.339 & 0.637\\       
   \bottomrule
\end{tabular}
}
\caption{ Imagenet $256\times256$ data set: image reconstruction (PSNR, SSIM) obtained by the compared methods. \textbf{Bold}: best, \underline{underline}: second.\label{tab:Imagenet-metrics}}
\end{table*}

\subsection{Compared methods \& figures-of-merit}
The proposed method has been compared to state-of-the-art methods related to the rationales motivating PnP-SGS:
\begin{itemize}
    \item SPA \cite{vono2019split}: split-and-augmented Gibbs sampler is an extension of SGS; in our experiments, it is used with a usual Tikhonov regularizer for deblurring and superresolution and with total-variation (TV) for inpainting;
    \item TV-ADMM: ADMM with a TV regularization;
    \item PnP-ADMM \cite{chan2016plug}: ADMM with a PnP regularization  chosen as DnCNN \cite{zhang2017beyond}; this can be interpreted as the deterministic counterpart of PnP-SGS;
     \item Score-SDE \cite{song2021scorebased}: implemented using the same DDPM as the one used by PnP-SGS;
    \item DDRM \cite{kawar2022denoising}: the denoising diffusion restoration model is implemented using the same DDPM as PnP-SGS;
    \item MCG \cite{chung2022improving}: manifold constrained gradients.   
\end{itemize} 
Note that PnP-ADMM, TV-ADMM, DDRM and Score-SDE  yield point estimates only. In contrast, PnP-SGS provides a comprehensive description of the targeted posterior distribution so that it permits to quantify uncertainties. It yields variances and credibility intervals and multiple statistics of the posterior for a variety of estimators such as MMSE and MAP. Implementation details are reported in Appendix \ref{sec:implem}.

The results are first qualitatively evaluated through visual inspection. Quantitative comparisons are conducted based on four widely-used metrics. The first two criteria are standard image reconstruction metrics, namely peak signal-to-noise-ratio (PSNR) and structural similarity index (SSIM). The two other criteria are perceptual metrics: Fréchet Inception Distance (FID), and Learned Perceptual Image Patch Similarity (LPIPS) distance. Results are averaged over $1000$ test images.

\subsection{Experimental results}

Tables \ref{tab:FFHQ-metrics} and \ref{tab:Imagenet-metrics} report the quantitative results in terms of image reconstruction and perceptual metrics for the two data sets FFHQ and Imagenet, respectively. The proposed method outperforms all the other compared methods by significant margins for the SNR and for the visual perception metrics. Particularly, DDRM and Score-SDE rely on a DDPM where the pre-trained generative model is exactly the same as the one implemented in PnP-SGS. Results appearing as N/A  correspond to tasks which are either not relevant for the model or not implemented by the original authors.

\begin{figure*}
    \centering
    \includegraphics[width=0.169\linewidth]{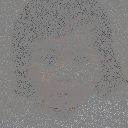}
    \hspace{-7px}
    \includegraphics[width=0.169\linewidth]{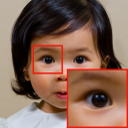}
    \hspace{-2px}
    \includegraphics[width=0.169\linewidth]{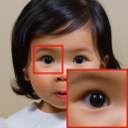}
    \hspace{-7px}
    \includegraphics[width=0.169\linewidth]{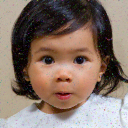} 
    \hspace{-7px}
    \includegraphics[width=0.169\linewidth]{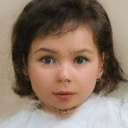} 
    \hspace{-7px}
    \includegraphics[width=0.169\linewidth]{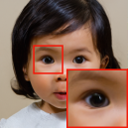}\\
    \vspace{-1px}
    \includegraphics[width=0.169\linewidth]{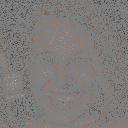}
    \hspace{-7px}
    \includegraphics[width=0.169\linewidth]{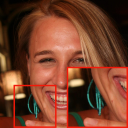}
    \hspace{-2px}
    \includegraphics[width=0.169\linewidth]{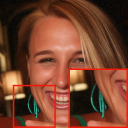}
   \hspace{-7px}
    \includegraphics[width=0.169\linewidth]{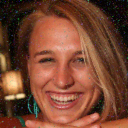} 
    \hspace{-7px}
    \includegraphics[width=0.169\linewidth]{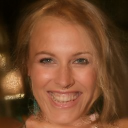} 
    \hspace{-7px}
    \includegraphics[width=0.169\linewidth]{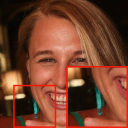}\\
    \vspace{-1px}
    \includegraphics[width=0.169\linewidth]{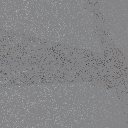}
    \hspace{-7px}
    \includegraphics[width=0.169\linewidth]{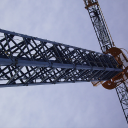}
    \hspace{-2px}
    \includegraphics[width=0.169\linewidth]{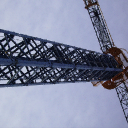}
    \hspace{-7px}
    \includegraphics[width=0.169\linewidth]{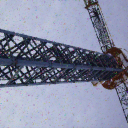} 
    \hspace{-7px}
    \includegraphics[width=0.169\linewidth]{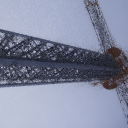} 
    \hspace{-7px}
    \includegraphics[width=0.169\linewidth]{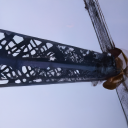}\\
    \vspace{-1px}
    \includegraphics[width=0.169\linewidth]{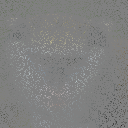}
    \hspace{-7px}
    \includegraphics[width=0.169\linewidth]{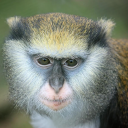}
    \hspace{-2px}
    \includegraphics[width=0.169\linewidth]{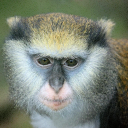}
    \hspace{-7px}
    \includegraphics[width=0.169\linewidth]{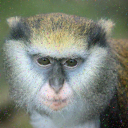} 
    \hspace{-7px}
    \includegraphics[width=0.169\linewidth]{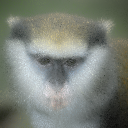} 
    \hspace{-7px}
    \includegraphics[width=0.169\linewidth]{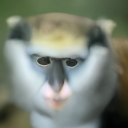}
    \caption{Inpainting task on the FFHQ (two top rows) and Imagenet data sets (two bottom rows), from left to right: measurement, true image, PnP-SGS, SPA \cite{vono2019split},  DDRM \cite{kawar2022denoising}, MCG \cite{chung2022improving}.}
    \label{fig:inpainting}
\end{figure*}

Fig. \ref{fig:inpainting} permits to assess the performances by visual inspection when inpainting 4 test images taken from the FFHQ and Imagenet data sets. In particular, PnP-SGS is compared to state-of-the-art methods which are known to be robust to measurement noise. PnP-SGS is able to provide high-quality reconstructions that are crisp and realistic. In particular it is able to recover more granular details.

\begin{figure}
    \centering
    \includegraphics[width=0.24\linewidth]{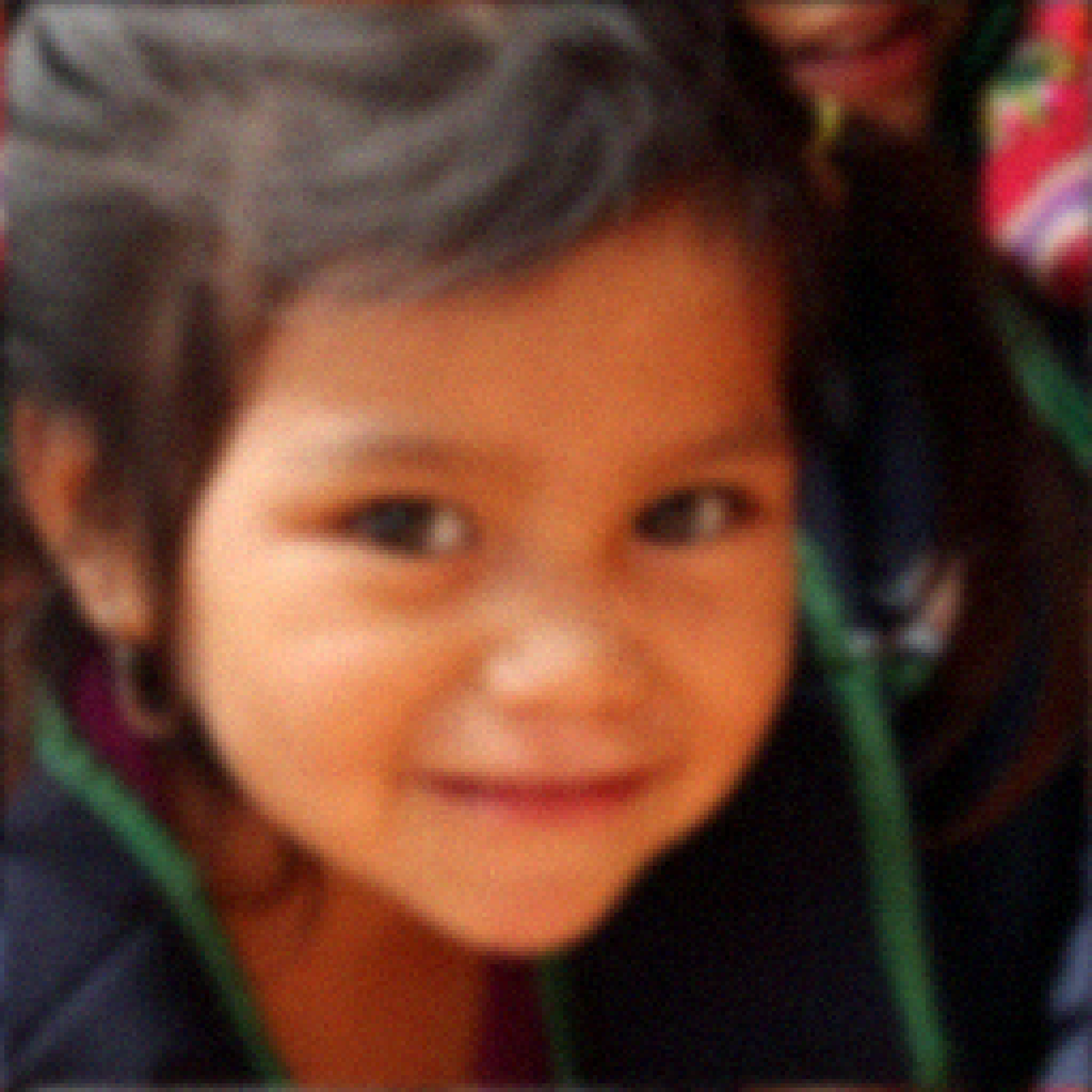}
    \hspace{-7px}
    \includegraphics[width=0.24\linewidth]{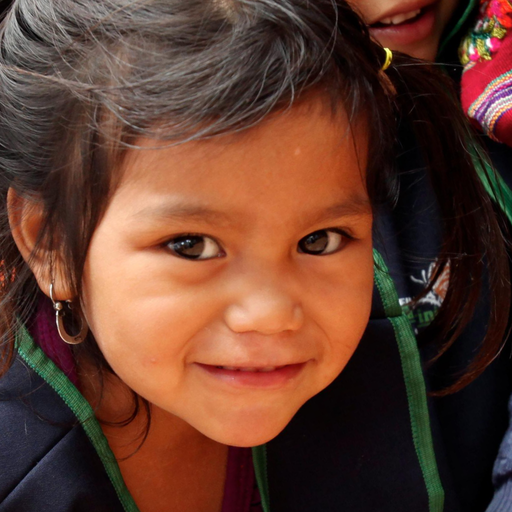}
    \hspace{-7px}
    \includegraphics[width=0.24\linewidth]{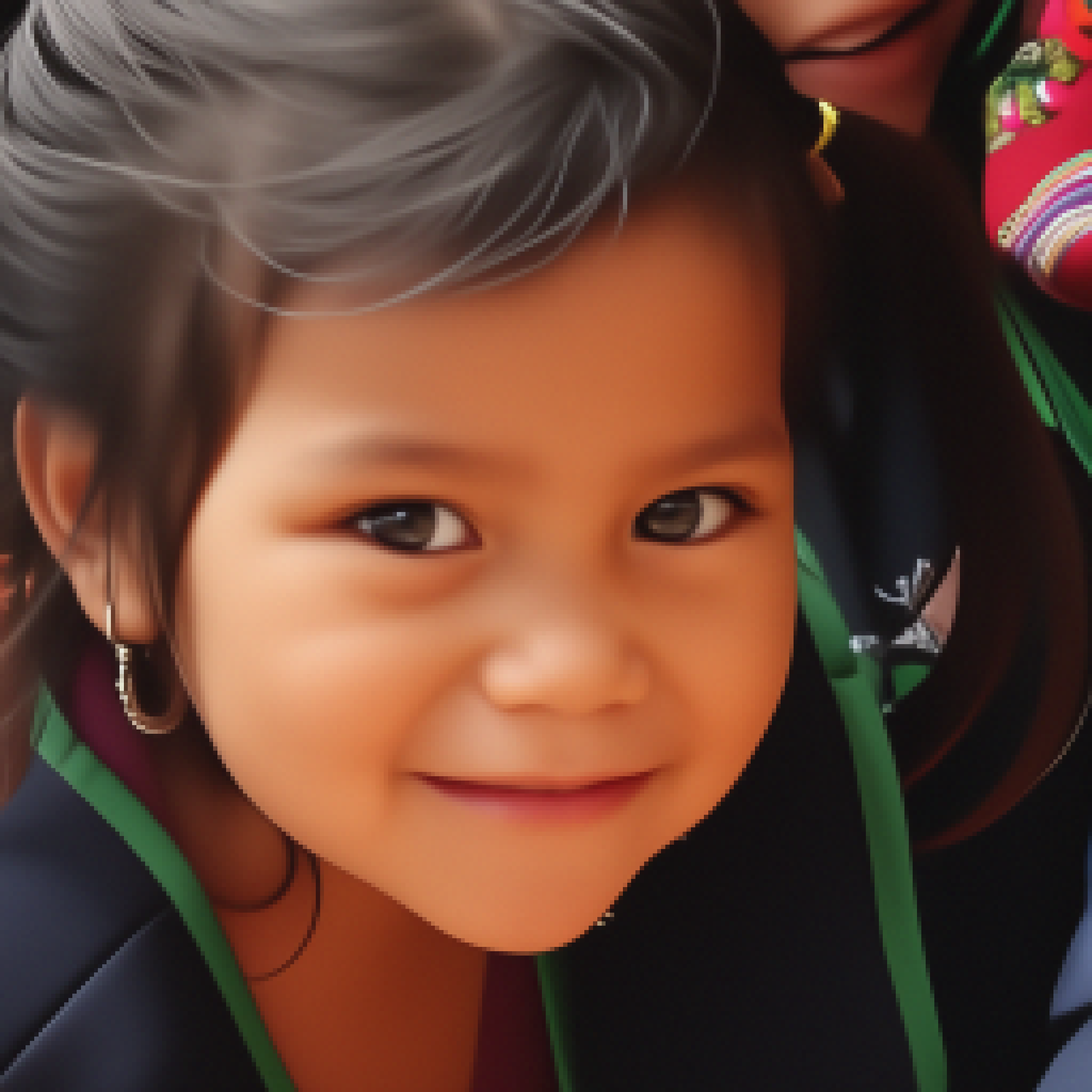}
    \hspace{-7px}
    \includegraphics[width=0.24\linewidth]{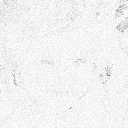}
    \\
    \vspace{-7px}
    \includegraphics[width=0.24\linewidth]{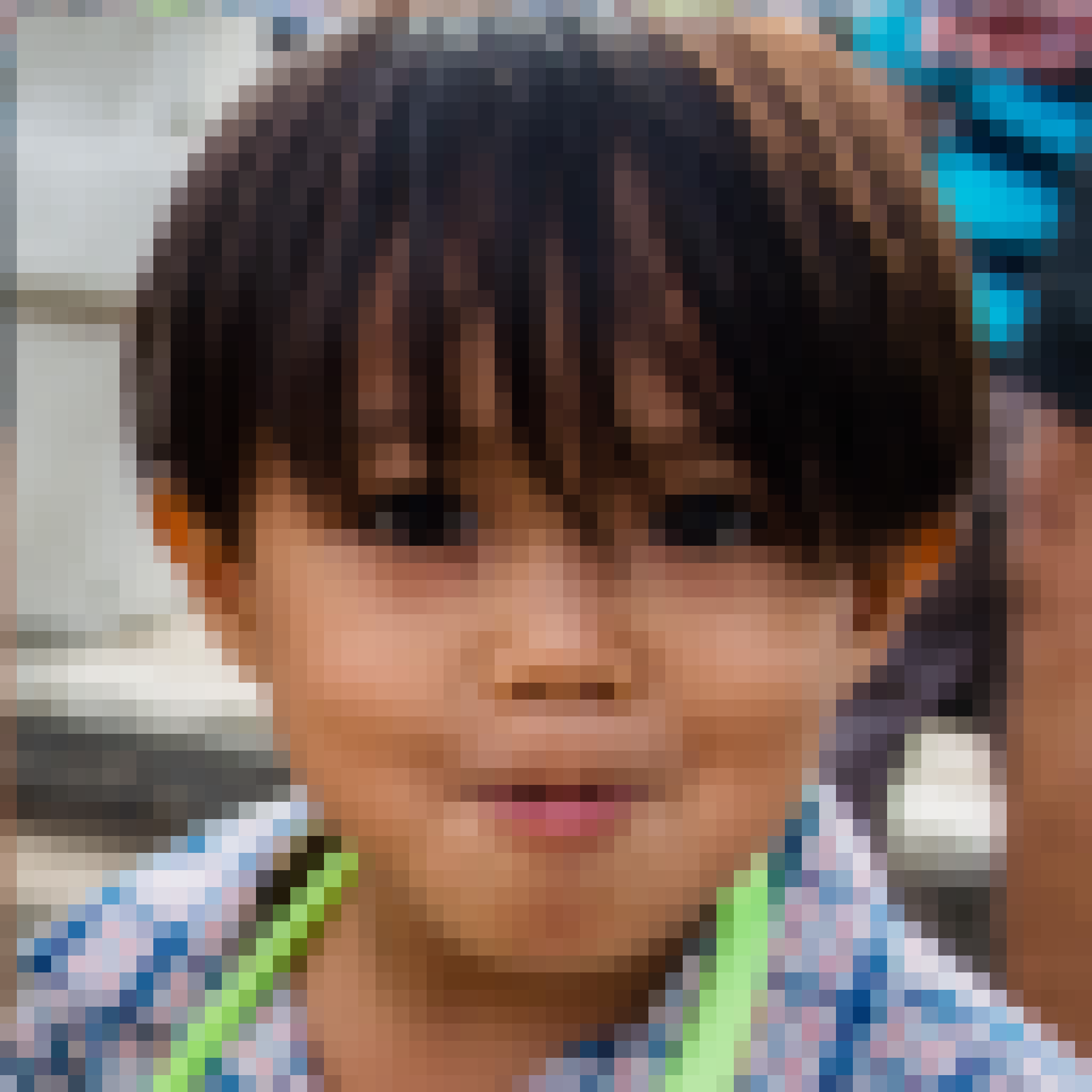}
    \hspace{-7px}
    \includegraphics[width=0.24\linewidth]{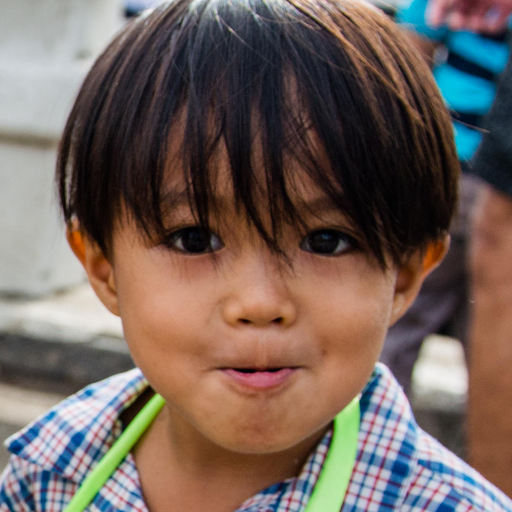}
    \hspace{-7px}
    \includegraphics[width=0.24\linewidth]{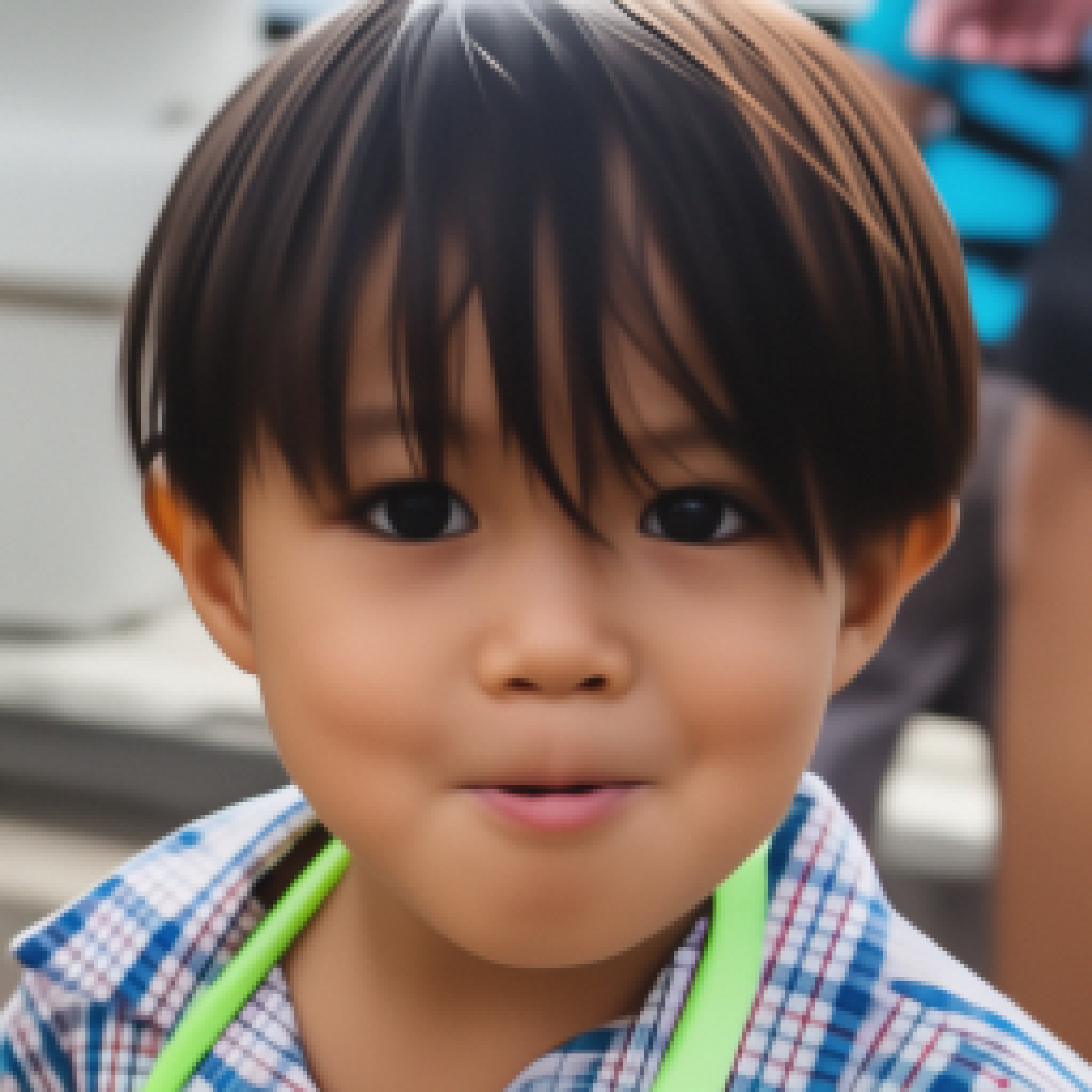} 
    \hspace{-7px}
    \includegraphics[width=0.24\linewidth]{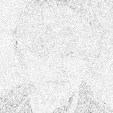}   
    \\
    \vspace{-7px}
    \includegraphics[width=0.24\linewidth]{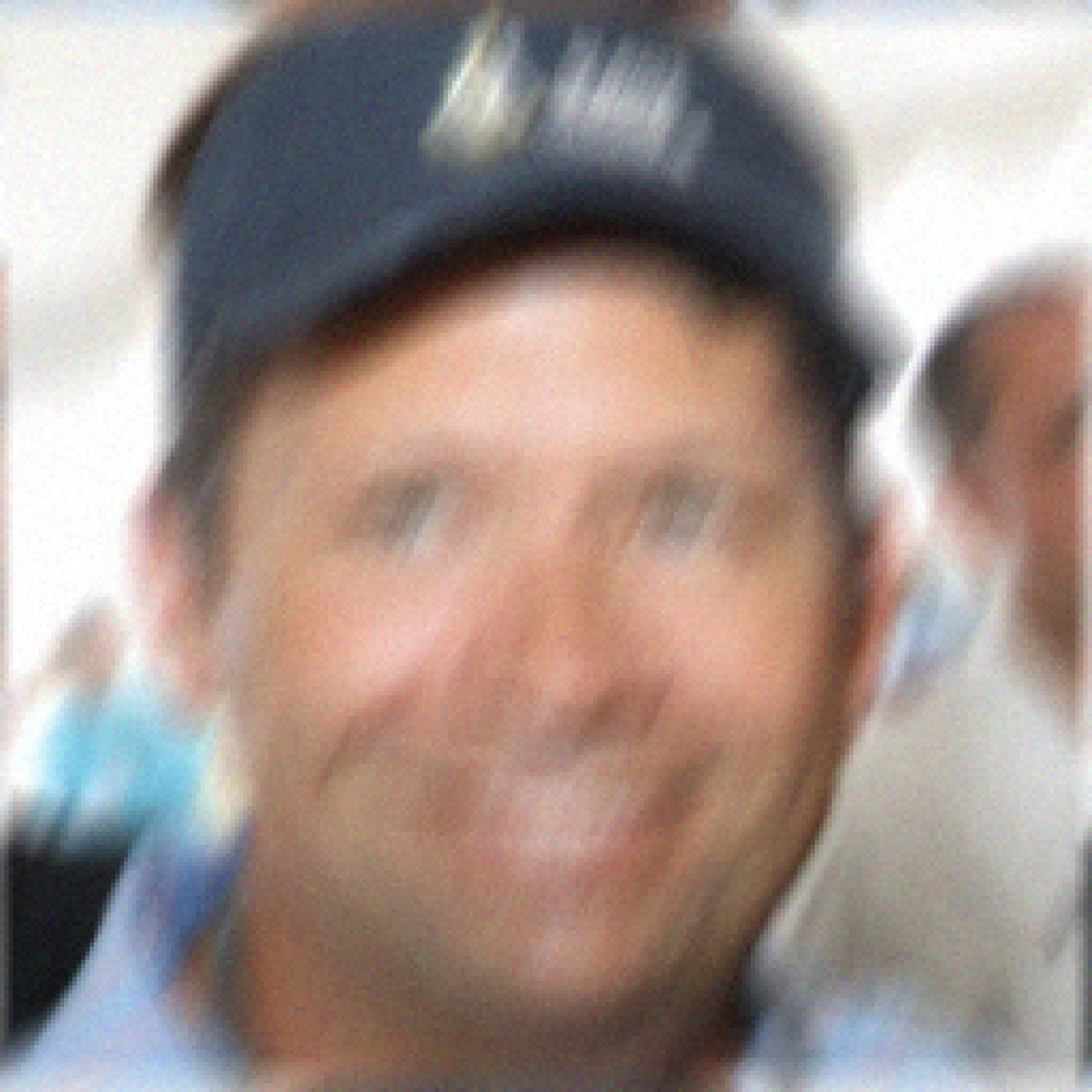}
    \hspace{-7px}
    \includegraphics[width=0.24\linewidth]{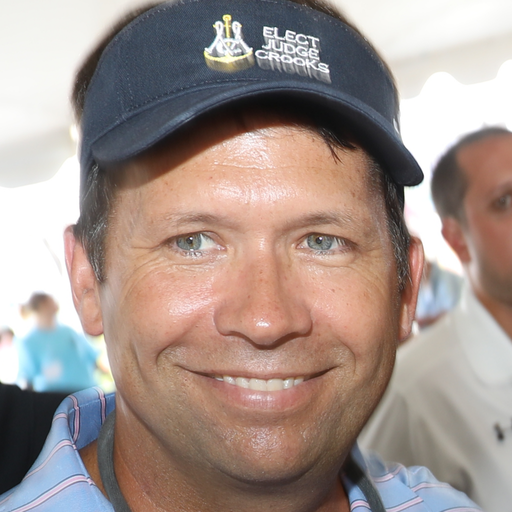}
    \hspace{-7px}
    \includegraphics[width=0.24\linewidth]{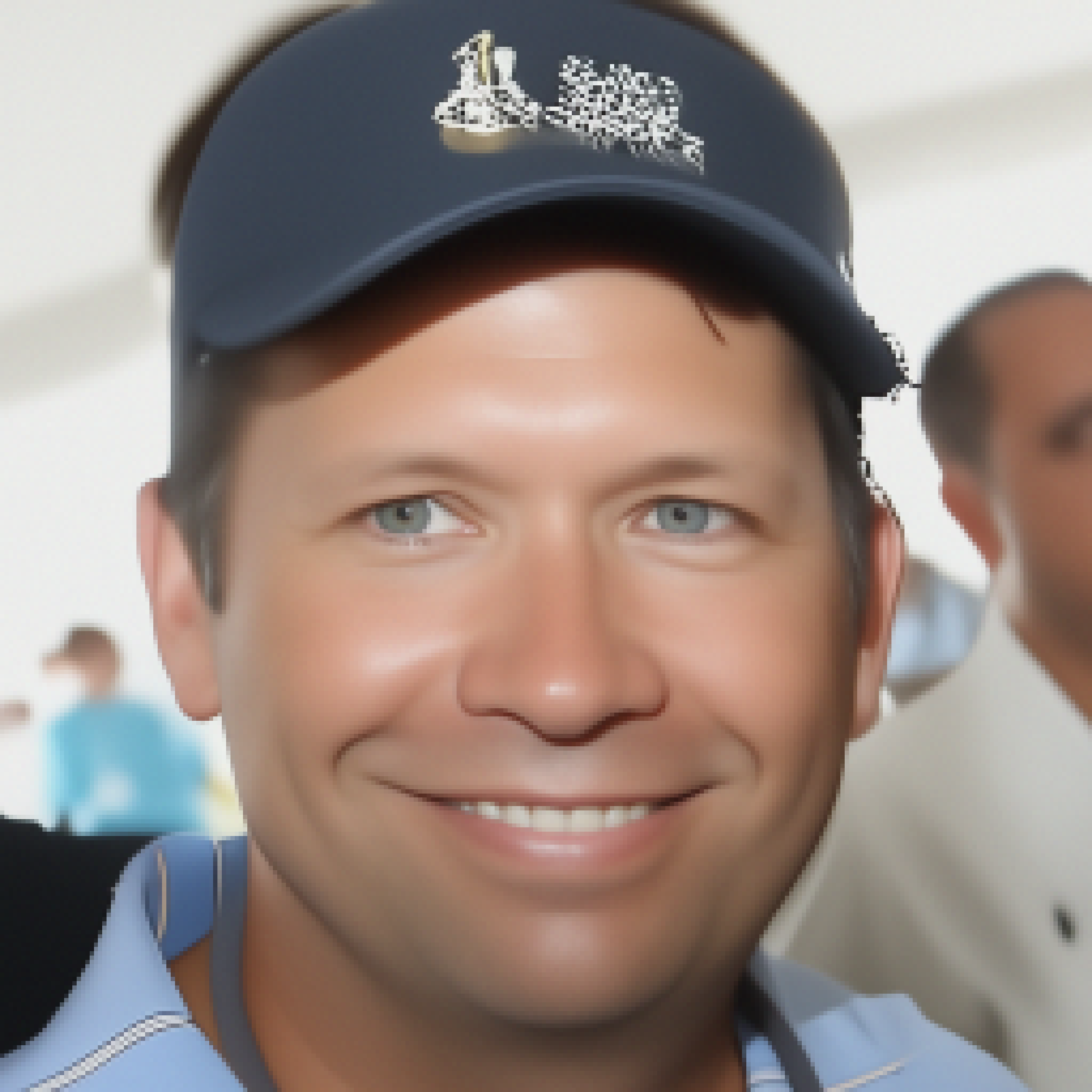} 
    \hspace{-7px}
    \includegraphics[width=0.24\linewidth]{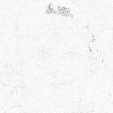}   
    \\
    \vspace{-7px}
    \includegraphics[width=0.24\linewidth]{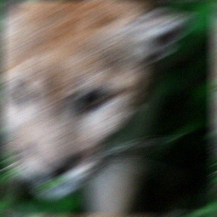}
    \hspace{-7px}
    \includegraphics[width=0.24\linewidth]{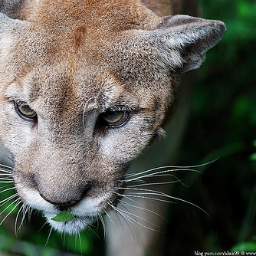}
    \hspace{-7px}
    \includegraphics[width=0.24\linewidth]{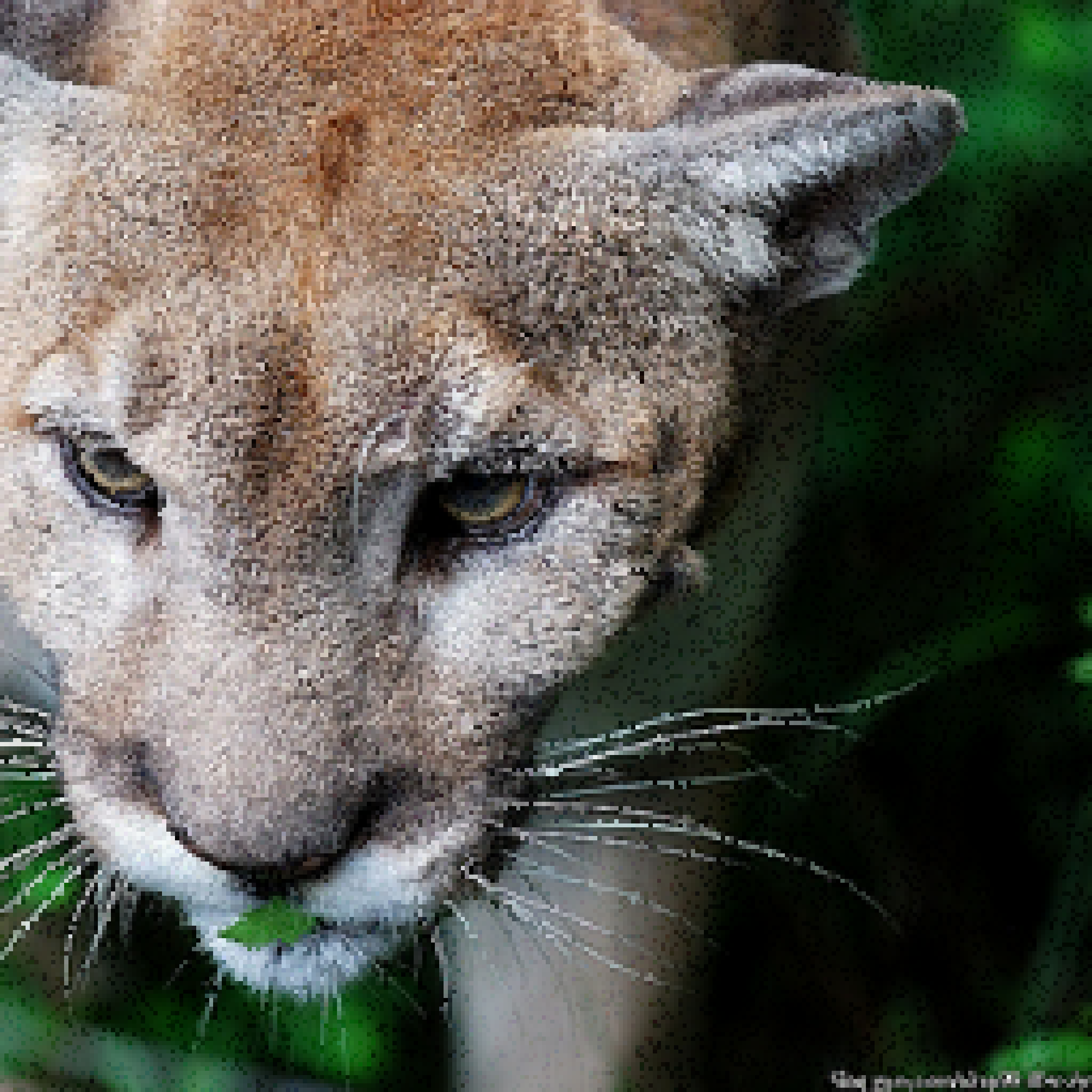} 
    \hspace{-7px}
    \includegraphics[width=0.24\linewidth]{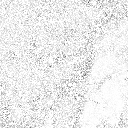}    
    
    \caption{From left to right: measurement, true image, MMSE estimate, pixel-wise 90\% credibility intervals. From top to bottom: Gaussian blur, motion blur, superresolution. The first three and the last images are from the FFHQ and Imagenet data sets, respectively.}
    \label{fig:blur_SR}
\end{figure}

\begin{figure*}
    \centering
    \includegraphics[scale=0.8]{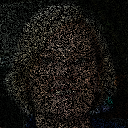} \hspace{-7px}
    \includegraphics[scale=0.8]{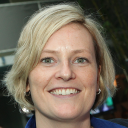}
    \includegraphics[scale=0.8]{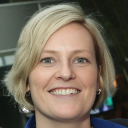} \hspace{-7px}
    \includegraphics[scale=0.8]{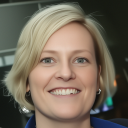} \hspace{-7px}
    \includegraphics[scale=0.8]{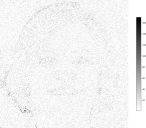}
    \includegraphics[scale=0.599]{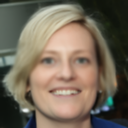} \hspace{-7px}
    \caption{Image inpainting, from left to right: measurement, true image, MMSE estimate $\hat{\bx}_{\mathrm{MMSE}}$, MMSE estimate $\hat{\bz}_{\mathrm{MMSE}}$, pixel-wise 90\% credibility intervals, averaged samples generated by MCG \cite{chung2022improving}}
    \label{fig:credib}
\end{figure*}

As already stated, the proposed PnP-SGS generate samples asymptotically distributed according to the posterior distribution. These samples can be used to approximate various Bayesian estimators but also to derive credibility intervals. %
Fig. \ref{fig:blur_SR} illustrates this advantage by depicting various restored images (in term of MMSE estimates) as well as 90\% credibility intervals for different tasks. This added value cannot be provided by optimization-based methods, e.g., TV-ADMM and PnP-ADMM, which provide point estimates only. Besides, stochastic samplers such as DDRM, MCG and Score-SDE are not able to provide this information either. Indeed, they do not generate multiple samples drawn from a stationary posterior. Several runs of these methods produce outputs that may be individually relevant but that are not consistent between them in their details, in particular because they originate from different noise realizations. This is also why averaging multiple outputs of these methods does not yield reliable MMSE estimators but rather tends to recover blurred images, as illustrated in Fig. \ref{fig:credib} ($6$th right panel for MCG).

When targeting \eqref{eq:spdist}, PnP-SGS generates two sets of samples $\left\{\bx^{(n)}\right\}_n$ and $\left\{\bz^{(n)}\right\}_n$ that are marginally distributed according to the marginals of $\pi_{\rho}(\bx,\bz)$. It follows a splitting strategy where the variables $\bx$ and $\bz$ are coupled thanks to a quadratic kernel that is controlled by the parameter $\rho$. Thus the posterior means $\hat{\bx}_{\mathrm{MMSE}}=\mathrm{E}[\bx|\by]$ and $\hat{\bz}_{\mathrm{MMSE}}=\mathrm{E}[\bz|\by]$ should be very similar up to some variations adjusted by the coupling parameter $\rho$. 
Fig. \ref{fig:credib} depicts the two estimates as well as pixel-wise 90\% credibility intervals. As expected, slight differences are observed. In particular, the point estimate $\hat{\bx}_{\mathrm{MMSE}}$ seems to be characterized by sharper details (better viewed by zooming on screen). Recall that this estimate is closer to the observation, while $\hat{\bz}_{\mathrm{MMSE}}$ is closer to the prior, therefore smoother. 

\begin{table}[]
\setlength{\tabcolsep}{6pt}
    \centering
    \begin{tabular}{ccccccc}
       \hline
        PnP-SGS  &  SPA & PnP-ADMM  & Score-SDE  &
        DDRM  & MCG \\
        \hline
        13.81    & 218.90      & 3.63     & 36.71     & 2.03       & 80.10 \\
        \hline
    \end{tabular}
    \caption{Inpainting: computational times (s.) of the compared methods.\label{tab:runtime}}
\end{table}

Table \ref{tab:runtime} reports the execution times for the task of inpainting of the various methods implemented on a single GTX 2080Ti GPU. 
Noticeably, the computational time of PnP-SGS  is similar to its competitors. In particular, this stochastic MCMC method (13.81s) is more than twice faster than Score-SDE (36.71s). It remains within a factor less than 4 with respect to PnP-ADMM (3.63s), its deterministic counterpart. 
The price to pay to get quantified uncertainties sounds very reasonable.
It is worth noting that using a DDPM-based PnP with SGS significantly reduces the number of iterations required by the sampler to reach the steady regime, which explains the reduced computational cost with respect to SPA.

\section{Conclusion}
\label{sec:conclusion}
This work proposes the plug-and-play split Gibbs sampler (PnP-SGS) as a stochastic counterpart of the well-known PnP-ADMM. 
Thanks to the SGS divide-to-conquer strategy, the PnP-SGS algorithm permits to target a posterior distribution that involves an implicit PnP prior where the regularization is ensured by some efficient stochastic denoiser. 
The proposed methodology can make use of any well-suited PnP prior, depending on the final application.
For instance, it can be based on a denoising diffusion probabilistic model (DDPM), as proposed here, since it appears that a DDPM can be turned into a Bayesian sampler of a denoising problem. 
With the same versatility as PnP-ADMM, sampling from the posterior distribution noticeably permits to build credibility intervals on top of point estimates. 
Extensive numerical experiments show that the proposed approach competes favourably with existing state-of-the-art models on typical imaging problems, namely deblurring, inpainting and superresolution. The quantitative performances are at least comparable when not better, while the computational times remain very moderate as well.
PnP-SGS appears as a scalable MCMC sampling method that can benefit from the most recent progress in machine (deep) learning at the price of a reasonable computational cost.

\newpage

\appendix

\section*{Appendices}
\addcontentsline{toc}{section}{Appendices}
\renewcommand{\thesubsection}{\Alph{subsection}}

\subsection{Estimating the noise level} \label{sec:denoise_exp}
The stochastic denoising task corresponding to the conditional distribution (\ref{eq:condz}) requires an estimation of the level $\hat{\sigma}=\Phi(\bx^{(n)})$ of a Gaussian noise assumed to affect the current state $\bx^{(n)}$ at each iteration of PnP-SGS (see Algo. \ref{alg:PnP-SGS}, line 5). The problem of estimating the level of the noise corrupting natural images has motivated plenty of research works, see  \cite{donoho1994ideal, guo2021gaussian, li2022single}. In our implementations, this estimation has been carried out following the strategy proposed in  \cite{donoho1994ideal}. This robust wavelet-based estimator  is already implemented in the library \texttt{scikit-image} (aka \texttt{skimage}) as the function  \texttt{estimate\_sigma()}. When handling RGB natural images, this function has been used with the parameter \texttt{average\_sigmas=True} to average the noise level estimates over the three channels.

\subsection{Inverting the variance function $\alpha(\cdot)$} \label{app:start_time}
Given the current estimate of the noise level $\hat{\sigma}$, sampling according to (\ref{eq:condz}) is achieved by performing the backward diffusion with kernel \eqref{eq:rev} from a time instant $\widehat{t^*}$ such that $\hat{\sigma}^2 = \alpha(\widehat{t^*})$ where $\alpha(t)$ is defined by \eqref{eq:def_alpha}. 
This diffusion scheduling function is controlled by the function $b(\cdot)$ that adjusts the variance of the forward transition kernel \eqref{eq:forw} from $t-1$ to $t$. Various choices of $b(\cdot)$ exist in the literature. We have tested two particular choices. 
For experiments with FFHQ, we chose a linearly increasing function 
\begin{equation}
    b(t) = b(0) + r t    
\end{equation}
where $b(0)=10^{-4}$ and the slope $r$ has been adjusted such that  $b(T) = 2.0 \times 10^{-2}$ \cite{ho2020denoising}.
For experiments with ImageNet, we adopted the cosine-based variance schedule  \cite{nichol2021improved} 
\begin{equation}
    \alpha(t)=1-\frac{\gamma(t)}{\gamma(0)}
\end{equation}
with $\gamma(t)=\cos \left(\frac{\pi}{2} \frac{t / T+s}{1+s} \right)^2$. 
In both cases, an explicit inverse function $\alpha^{-1}(\cdot)$ can be derived, which yields ${\widehat{t^*}} = \alpha^{-1}(\hat{\sigma}^2)$. 
For more complex scheduling functions, an alternative is to use a tabbing strategy, which saves computation cost as well. Given a pre-computed list of $T+1$ values $\boldsymbol{\alpha}=\left\{\alpha(0),\ldots,\alpha(T)\right\}$, the diffusion start time is set to 
$${\widehat{t^*}} = \operatornamewithlimits{argmin}_{t\in \left\{0,\ldots,T\right\}} |\alpha(t) - \hat{\sigma}^2|.$$
In our experiments, the scheduling functions have been sampled on $T=1000$ regularly spaced time instants.

\subsection{Experimental details} \label{sec:implem}

\subsubsection{Proposed PnP-SGS} 

\begin{figure}
\centering
\includegraphics[scale=0.45]{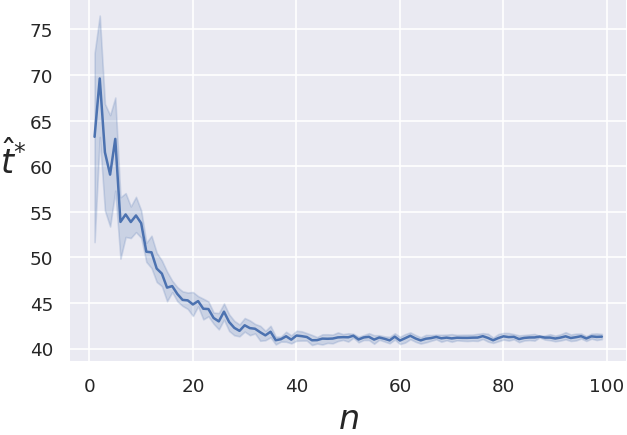}
\caption{Inpainting: evolution of ${\widehat{t^*}}$ along the PnP-SGS iterations ($T=1000$). Results have been averaged over $100$ runs conducted on the same image. Shaded areas stand for the corresponding standard deviation.}
\label{fig:t_star}
\end{figure}

For the experiments on the FFHQ data set, the pre-trained DDPM has been taken from \cite{choi2021ilvr} also available online\footnote{\url{https://github.com/jychoi118/ilvr\_adm}} and the coupling parameter has been manually set to $ \rho = 0.7$. For the Imagenet data set, we have used the pre-trained DDPM of \cite{dhariwal2021diffusion} and available online\footnote{\url{https://github.com/openai/guided-diffusion}} and the coupling parameter has been fixed as $\rho=1.625$.
For all experiments, the number of iterations of the PnP-SGS has been fixed as $N_{\mathrm{MC}}=100$ including $N_{\mathrm{bi}}=20$ burn-in iterations.
In Section \ref{sec:experiments}, the estimated number $\widehat{t^*}$ of denoising steps is automatically adjusted by the procedure described in Section \ref{subsec:insights_t}. At the first iteration of the PnP-SGS, $\widehat{t^*}$ is usually a fraction of $T$. Along the iterations of the PnP-SGS, this number reduces and then stabilizes around a small fraction of $T$, as illustrated in Fig. \ref{fig:t_star} where the initial value of $\widehat{t^*}$ is around $0.07T$ for the inpainting task.

During the burn-in period, instead of applying the kernel \eqref{eq:rev} for $t=\widehat{t^*},\ldots,1$, we suspend the process in the middle of the diffusion, i.e., $t=\widehat{t^*},\ldots,\frac{\widehat{t^*}}{2}$. This early-stopping trick not only provides empirically better results but also allows the computational burden to be lightened by reducing the number of DNN evaluations.

\subsubsection{Compared methods } 

DDRM, MCG and Score-SDE are implemented using the same pre-trained model as PnP-SGS (see above). Additional details are listed below:

\begin{itemize}
    \item DDRM: all experiments have been performed with the default setting $\eta_B=1.0$  and $\eta=0.85$. For the Gaussian deblurring task, the forward model was implemented by separable 1D convolutions for efficient SVD.
    \item MCG: the variance scheduling function  $\alpha(\cdot)$  has been chosen as the one used by PnP-SGS. At each step, complementary data consistency steps are applied as Euclidean projections onto the measurement set $\mathcal{C}=\left\{\bx_i \mid \bH \bx_i=\by_i, \by_i \sim p\left(\by_i \mid \by_0\right)\right\}$ 
    \item Score-SDE solves the inverse problems by iteratively applying a denoising step followed by data consistency projections onto the measurement set $\mathcal{C}$, as in MCG.
    \item  PnP-ADMM: the implementation is from the \texttt{SCICO}\footnote{`\url{https://scico.readthedocs.io}}. library. The parameters are set to $\rho=0.2$ (ADMM penalty parameter) and $\texttt{maxiter}=12$. Proximal mappings use the pretrained DnCNN denoiser \cite{zhang2017beyond}.
    \item TV-ADMM uses the isotropic regularization. The regularization parameter $\lambda$ and some penalty parameter $\rho$ linked to the splitting have been adjusted by grid search to reach the best performance. Final values are  $(\lambda, \rho)=(2.7 \times 10^{-2},1.4\times 10^{1})$ for deblurring, $(\lambda, \rho)=(2.7\times 10^{-2}, 1.0 \times 10^{-2})$ for inpainting and $(\lambda, \rho_1,\rho_2)=(2.7\times 10^{-2}, 1.0 \times 10^{-2})$ for superresolution which requires a double splitting.
\end{itemize}

\bibliographystyle{IEEEtran}
\bibliography{strings_all_ref,reference}

\vfill

\end{document}